\documentclass[journal,onecolumn]{article}

\usepackage[a4paper,bindingoffset=0.2in,%
            left=1in,right=1in,top=1in,bottom=1in,%
            footskip=.25in]{geometry}
\usepackage{blindtext}

\usepackage{epsfig,color,hl_short,amsfonts,amsmath}
\usepackage[lined,linesnumbered,boxed]{algorithm2e}
\usepackage{tikz}
\usepackage{enumitem}
\usepackage[draft]{fixme}
\usepackage{setspace}
\usepackage{dsfont}
\usepackage{float}
\usepackage{cite}
\usepackage{color}
\usepackage{graphicx}
\usepackage{textcomp}
\definecolor{listinggray}{gray}{0.9}
\definecolor{lbcolor}{rgb}{0.9,0.9,0.9}

\definecolor{pblue}{rgb}{0.1,0.1,1}
\definecolor{pgreen}{rgb}{0,0.5,0}
\definecolor{pred}{rgb}{0.9,0,0}
\definecolor{pgrey}{rgb}{0.46,0.45,0.48}

\usepackage{listings}
\usepackage{caption}

\DeclareCaptionFont{black}{ \color{black} } \DeclareCaptionFormat{listing}{
 \colorbox[cmyk]{0, 0, 0,0}{
  \parbox{\textwidth}{\hspace{0pt}#1#2#3\vspace{0pt}}
 }
} 

\nc{\varApprox}[1]{\ensuremath{q\ifthenelse{\equal{#1}{}}{}{\left(#1\right)}}}
\nc{\varApproxPart}[2]{\ensuremath{\varApprox{}_{#1}\ifthenelse{\equal{#2}{}}{}{\left(#2\right)}}}
\nc{\varEnt}{\ensuremath{\calF(\varApprox{})}\xspace}
\nc{\entropy}{\ensuremath{\calH}}

\nc{\noObs}{\ensuremath{\#\text{Obs}}}
\nc{\const}{\text{ const.}}
\nc{\Identity}{\ensuremath{\boldsymbol{\mathrm{I}}}}
\nc{\GenMF}{generalized mean-field\xspace}
\nc{\Exp}{\ensuremath{\mathbb{E}}}
\DeclareMathOperator*{\argmax}{arg\,max}

\DeclareTextFontCommand{\mytexttt}{\ttfamily\hyphenchar\font=45\relax}

\urldef{\webOpenMP}\url{http://users.eecs.northwestern.edu/~ewkliao/Kmeans/index.html}
\urldef{\webMurphy}\url{http://www.cs.ubc.ca/~emurphyk/Software/bnsoft.html}

\newcommand{\drop}[1]{}

\begin{document}

\title{Probabilistic Graphical Models on Multi-Core CPUs \\ using Java 8}

\author{\textbf{Andr\'{e}s R. Masegosa}\thanks{Corresponding author: Andr\'{e}s R. Masegosa (Email: andres.masegosa@idi.ntnu.no).}, \\ Department of Computer and Information Science, \\Norwegian University of Science and Technology, \\ Trondheim, Norway \\\\ \textbf{Ana M. Mart\'{\i}nez} and \textbf{Hanen Borchani},\\ Department of Computer Science, \\ Aalborg University, Denmark}


\maketitle

\doublespacing

\begin{abstract}
In this paper, we discuss software design issues related to the development of parallel computational intelligence algorithms on multi-core CPUs, using the new Java 8 functional programming features. In particular, we focus on probabilistic graphical models (PGMs) and present the parallelisation of a collection of algorithms that deal with inference and learning of PGMs from data. Namely, maximum likelihood estimation, importance sampling, and greedy search for solving combinatorial optimisation problems. Through these concrete examples, we tackle the problem of defining efficient data structures for PGMs and parallel processing of same-size batches of data sets using Java 8 features. We also provide straightforward techniques to code parallel algorithms that seamlessly exploit multi-core processors. The experimental analysis, carried out using our open source AMIDST (Analysis of MassIve Data STreams) Java toolbox, shows the merits of the proposed solutions.
\end{abstract}



\section{Introduction}\label{sec:introduction}

In the last decade, the chip-manufacturing industry has relied on multi-core architectures to boost the performance of computing processors. Previously, improvements of computer power were driven by increasing the operating frequency of the chips. However, this frequency scaling strategy started to produce diminishing gains due to several factors. For example, the power consumption exponentially increases with each factorial increase of the operating frequency (that is, simply because power is consumed by the CMOS circuits every time they change their state). This problem cannot be mitigated by shrinking the components of the processors due to leakage problems \cite{frank2002power,gelsinger2001microprocessors}. As a result of using several processing cores on a chip, industry was able to multiply the computing performance of the processors while keeping the clock frequency fixed.

Computational intelligence methods have been strongly influenced by the emergence of parallel computing architectures. The literature on distributed and parallel algorithms has exploded in the last years \cite{liu2006distributed,chu2007map,umbarkar2013review,robles2009evolutionary,
luque2005parallel,adeli2006cost}. Consequently, computational intelligence software libraries, especially in the field of ``Big data'', have emerged to help developers to leverage the computational power of systems with multiple processing units \cite{JMLR:v15:agarwal14a,JMLR:v16:morales15a}. Classic scientific programming libraries like R or Matlab have also tried to adapt to multi-core CPUs or computer clusters by providing specialised libraries \cite{ParallelR,sharma2009matlab,Kane:Emerson:Weston:2013:JSSOBK}. 

However, programming computational intelligence methods on parallel architectures is not an easy task yet. Apart from the latter developments, there are many other parallel programming languages such as Orca \cite{ORCA:WCMS81}, Occam ABCL \cite{May:1983:OCCAM}, SNOW \cite{snowRPackage}, MPI \cite{Forum:1994:MPI}, PARLOG \cite{parlog1987}, and Scala \cite{scala-overview-tech-report}. Nevertheless, at the time of writing, none of them seems to be widely accepted by the computational intelligence community. For example, if we take a look at some reference venues such as the JMLR Software track\footnote{\url{http://www.jmlr.org/mloss/}}, IEEE Fuzzy Software for Soft Computing series \cite{alcala2013special}, and the Journal of Statistical Software\footnote{\url{http://www.jstatsoft.org/}}, the vast majority of the proposed software libraries do not rely on any of these parallel programming languages\cite{AlcalaFdezAlonso15}. Moreover, only a few released libraries contain implementations of computational intelligence techniques that are able to exploit multi-core CPUs\cite{piccolo2012ml,schmidberger2009state,graphLab2010,petuum2015}. Enterprise-supported scientific computing libraries constitute an exception \cite{sharma2009matlab}, because in general programming concurrent and parallel applications is tough, time consuming, error prone, and requires specific programming skills that may be beyond the expertise of C. I. researchers. 

The latest Java release (JDK 8.0, March 2014) is an attempt to bring a new API to code transparent and seamless parallel applications to the general public (that is, non-specialized programmers in parallel and concurrent methods) \cite{warburton2014java}. Note that, from the very beginning, Java architects tried to address the problem of programming parallel and concurrent applications. Java 1.0 had threads and locks, Java 5 included thread pools and concurrent collections, while Java 7 added the fork/join framework. All these functionalities made parallel programming more practical, yet not trivial to accomplish \cite{urma2014java}. To handle this, Java 8 introduces a completely new approach to parallelism based on a functional programming style \cite{warburton2014java} through the use of map-reduce operations \cite{Dean2004mapReduce}. Although map-reduce operations existed already as higher order functions in the programming literature, they only started to be widely used with the rise of parallel processing \cite{Dean2004mapReduce}. Roughly speaking, map-reduce operations are usually employed for parallel processing of collections of elements (\texttt{Stream} objects in Java 8). The \texttt{map} operation applies a stateless user-defined transformer function (mapper) to each element of the collection, whereas the \texttt{reduce} operation applies another stateless user-defined function (reducer) that combines the outputs of the mapper to produce a single output result. Both the mapper and the reducer can be concisely described using \texttt{lambda} expressions in Java 8. The stateless nature of the mapper and the reducer allows to safely parallelise their application over the different elements of the collection. Note that the term MapReduce commonly refers to two different concepts: 1) the Google private parallel engine (out of the scope of this paper); and 2) the MapReduce programming paradigm, which corresponds to the description above and commonly deals with problems such as distributed data or serialization \cite{Dean2004mapReduce}. In this work we consider the map-reduce concept from a functional programming perspective. As we focus on multi-core instances we do not need to deal with issues related to distributed systems.

Many experts argue that the introduction of these new functional programming features (that is, lambda expressions, map-reduce operations, and streams) is, in many ways, the most profound change to Java during the last few years \cite{urma2014java, warburton2014java}.

As a concrete example of the potential of the new Java 8 API, let us assume that we have a collection of documents stored in a \verb+List<String>+ object, called docs, and we want to produce, exploiting a multi-core CPU, a \texttt{Map} object containing all distinct words along with the number of occurrences of each word in the different documents. As shown in Algorithm 1, this task can be done using just three lines of code in Java 8. On the opposite, using the previous Java versions (or many other programming languages), coding this task would require dozens of lines of code as well as an explicit and complex management of multiple threads and concurrency. Nevertheless, using Java 8 presents the challenge of dealing with a new API and a new programming syntax that requires a learning curve. The syntax in Program 1 includes lambda expressions (such as \texttt{s -> s}) and map (\texttt{flatMap}) and reduce (\texttt{collect}) operations that will be later defined in Section \ref{sec:java8}. Readers can also find more details about the basics of Java 8 and more advanced functionalities in \cite{urma2014java}. 

\begin{lstlisting}[title= Program 1 - Word occurrences in documents, label = Prog1] 
Map<String, Long> wordToCount =  docs.parallelStream()
				   	      .flatMap(doc -> Stream.of(doc.split("\\s+")))
				  	      .collect(Collectors.toMap(s -> s, s -> 1, Integer::sum));

\end{lstlisting}
\vspace{0.1in}

Moreover, readers that are familiar with big data platforms such as Apache Spark\footnote{Apache Spark: \url{http://spark.apache.org}} \cite{Zaharia2010} or Apache Flink\footnote{Apache Flink: \url{https://flink.apache.org}} \cite{FLINK2015} will notice the syntax similarities between these two platforms and Java 8. This indeed may facilitate knowledge transfer among the mentioned platforms. However, it is important to note that their purposes are different. In fact, Java 8 streams target multi-core platforms, whereas Spark and Flink are intended to be used for distributed data processing on multi-node clusters. Spark and Flink could also be used for multi-core platforms but the main problem is that they may introduce overheads, which consequently discourages their use in such settings, as will be empirically demonstrated in Section \ref{ss:java8vsspark}. It is also worth highlighting that Java is quite a mature programming language (almost 20 years since the first release), widely used (around 3 billions devices), and has a large community (around 9 millions developers). Although Flink and Spark showed recently a great potential, Java 8 still entails a much more easy-to-use and robust solution for parallel processing in a multi-core environment. In addition, luckily, all these platforms can be used together since Java 8 can be incorporated on top of Spark or Flink to reduce potential overheads in terms of multi-core parallelisation\footnote{Last releases of Apache Flink and Apache Spark start to partially support Java 8.}.

Probabilistic graphical models (PGMs) \cite{Pearl88,lauritzen1996graphical,JensenNielsen07,koller2009probabilistic} have been state-of-the-art tools for reasoning under uncertainty in the last two decades. They have been successfully used in many real-world problems involving high number of variables with complex dependency structures. Application examples include text classification, genomics, automatic robot control, fault diagnosis, etc. (see \cite{pourret2008bayesian} for an extensive review on real applications).  Meanwhile, a large set of specialised libraries on PGMs based on several programming languages\footnote{See \url{http://www.cs.ubc.ca/~murphyk/Software/bnsoft.html} for a non-exhaustive list referencing around seventy different PGM libraries.} have also emerged in recent years from many different universities. However, to the best of our knowledge, none of them provides explicit and general support for parallel/concurrent processing. In our opinion, the main reasons for this absence are the relatively recent interest in concurrent/parallel algorithms as well as the complexity of coding such algorithms. 

In this paper we present some parts of the design of a software tool for PGMs and show how this design leverages some of the new Java 8 features for developing easy-to-code parallel algorithms on multi-core CPUs. In particular, we discuss the design of a collection of algorithms that are mainly related to the scalable representation, inference, and learning of PGMs from data. It is important here to highlight that the discussed software design solutions are generic and not restricted to the scope of PGMs, since they could be applied to other computational intelligence software libraries. Moreover, these solutions have been adopted in the recent development of our AMIDST (Analysis of MassIve Data STreams) Java toolbox\footnote{AMIDST is an open source toolbox written in Java and is available at \url{http://amidst.github.io/toolbox/} under the Apache Software License 2.0.}. It is important to clarify that the aim of this paper is not to demonstrate that our AMIDST package is more efficient than other available PGM softwares, but rather to show how convenient it is to use Java 8 for coding algorithms running on multi-core CPUs. More specifically, this paper aims to help other developers to understand the new Java 8 features and to reduce the learning curve by showing how Java 8 can solve concrete issues that arise when coding parallel/concurrent computational intelligence algorithms.

The remainder of this paper is organised as follows: In Section \ref{sec:background} we introduce the main concepts related to both Java 8 and PGMs. Section \ref{sec:dataStructures} describes some issues related to the design of data structures that support functional programming style while Section \ref{sec:algorithms} examines the parallelisation of algorithms for performing maximum likelihood estimation, Monte Carlo inference, and greedy search. Next, Section \ref{sec:dataStreams} discusses the parallel processing of data sets. Finally, we present a brief experimental evaluation showing the advantages of using the proposed design solutions for parallel computation in Section \ref{sec:experimentalEvaluation} and we conclude in Section \ref{sec:conclusion}. All the Java source codes of the provided examples can be downloaded from the AMIDST toolbox website \url{http://amidst.github.io/toolbox/}. 



\section{Background}\label{sec:background}

In this section, we start by briefly defining the main features related to the functional programming style in Java 8. Afterwards, we introduce Bayesian network models.

\subsection{Functional programming style in Java 8}\label{sec:java8}

Java 8 introduces a new way of programming based on method references, lambda expressions, streams, and parallel streams. In what follows, we provide a brief description of these different concepts and illustrate them with running examples. We also refer the readers to \cite{urma2014java} for a deep revision of all these concepts.

Consider that we want to determine the number of even and odd numbers in a given list of integers. Using prior versions of Java, we might consider to create a class \texttt{MyIntegerUtils} (see Program 2.1). This class defines an inner interface and three static methods, namely, \texttt{isEven}, \texttt{isOdd}, that take an input integer, and \texttt{filterNumberList}, that takes as inputs a list of integers and a predicate. This predicate might be later defined as an anonymous class implementing the interface \texttt{NbrPredicate}, as shown in Program 2.2.

In the following subsections, we will describe how Java 8, using its functional programming features, gets rid of all this verbose code and provides instead a much simpler and shorter syntax.

\subsubsection{\textbf{Method references}}\label{sec:java8:MethodReferences}

\textit{Method references} allow us to use an available method of a class and pass it directly as an argument to another method through the \verb+::+ syntax. Thus, the above example can be then simplified using the method references \verb+::isEven+ and \verb+::isOdd+ as shown in Program 2.3. By using method references, there is not a need to create any anonymous classes as previously done in Program 2.2. 

In this example, the interface \verb+NbrPredicate+ defines how the passed methods should look like. That is, it states that the passed methods accept an integer as a unique input argument and return a boolean value. Both methods \verb+isEven+ and \verb+isOdd+ from the \verb+MyIntegerUtils+ class fulfil this condition. 

\vspace{0.1in}
\begin{lstlisting}[title= Program 2.1 - A class to determine the number of even and odd numbers in a list of integers (using older versions of Java), label = Prog21] 
public class MyIntegerUtils { 
	public interface NbrPredicate{
		boolean test(Integer integer);
	}
	public static boolean isEven(Integer i){
		return i % 2 == 0; 
	}
	public static boolean isOdd(Integer i){
		return i % 2 == 1;
	}
	public static int filterList(List<Integer> input, NbrPredicate predicate){
		int count = 0;
		for (Integer i : input){
			if (predicate.test(i))
				count++;
		}
		return count;
	}
}
\end{lstlisting}
\vspace{0.1in}

\begin{lstlisting}[title= Program 2.2 - Anonymous classes implementing the interface NbrPredicate and an example of use, label = Prog22] 
public static void main (String[] args){
	NbrPredicate preEven = new NbrPredicate(){
		@Override
		public boolean test(Integer integer){return MyIntegerUtils.isEven(integer);}
	}
	NbrPredicate preOdd = new NbrPredicate(){
		@Override
		public boolean test(Integer integer){return MyIntegerUtils.isOdd(integer);}
	}
	
	List<Integer> list= Arrays.asList(0, 3, 5, 6, 8, 10, 12, 15);
	System.out.println("The count of even numbers is " +  MyIntegerUtils.filterList(list, preEven));
	System.out.println("The count of odd numbers is " + MyIntegerUtils.filterList(list, preOdd));
}	
\end{lstlisting}
\vspace{0.5in}

\begin{lstlisting}[title= Program 2.3 - Use of method references in Java 8, label = Prog23] 
List<Integer> list= Arrays.asList(0,3,5,6,8,10,12,15);
System.out.println("The count of even numbers is" + 
						MyIntegerUtils.filterList(list, MyIntegerUtils::isEven));
System.out.println("The count of odd numbers is" + 
						MyIntegerUtils.filterList(list, MyIntegerUtils::isOdd));
\end{lstlisting}
\vspace{0.1in}

\subsubsection{\textbf{Lambda expressions}}\label{sec:java8:lambda}

As an alternative to pass method references, Java 8 allows a richer use of functions as values by including lambda expressions (or anonymous functions). An example of a lambda expression could be  \verb'(int n) -> n + 1' and it refers to ``a function, that when called with the integer argument \verb'n', returns the value \verb'n+1'''. Although this kind of  functionality can of course be obtained by defining new methods and classes, in many cases it allows for concise statements which make the code much more simple and readable. Program 2.4 shows how our previous example looks like when using lambda expressions such as \verb'i -> i % 2 == 0', which defines a function that when called with the integer argument \verb'i', returns \verb'true' if the number is even or \verb'false' otherwise.  As can be seen, with this simple code we can get rid of of the previously defined statics methods \texttt{isEven} and \texttt{isOdd} of the \texttt{MyIntgerUtils} class.

Let us note that both lambda expressions in Program 2.4 are consistent with the \verb+NbrPredicate+ interface. Otherwise, we would get a compilation error (we also have type safety for lambda expressions). 

\begin{lstlisting}[title= Program 2.4 - Use of lambda expressions in Java 8, label = Prog24] 
List<Integer> list= Arrays.asList(0,3,5,6,8,10,12,15);
System.out.println("The count of even numbers is" + MyIntegerUtils.filterList(list, i -> i % 2 == 0));
System.out.println("The count of odd numbers is" + MyIntegerUtils.filterList(list, i -> i % 2 == 1));
\end{lstlisting}
\vspace{0.1in}

\subsubsection{\textbf{Streams, intermediate and terminal operations}}\label{sec:java8:streams}

A \texttt{java.util.Stream} represents a sequence of elements on which one or more operations can be performed. These operations are defined by the corresponding operators which are either: \textit{intermediate}, that return a new transformed stream (multiple intermediate operators can be sequentially applied such as \texttt{filter}, \texttt{map}, \texttt{skip}, and \texttt{limit}), or \textit{terminal}, that end the process of the sequence of elements by applying some final operation (for example \texttt{forEach}, \texttt{findFirst}, \texttt{collect}, and \texttt{count}). In terms of \emph{map-reduce} operations, intermediate operations can be regarded as \emph{map} operations whereas terminal operations can be matched to \emph{reduce} operations. 

In addition, all the Java collections (lists, sets, maps, etc.) contain the method \texttt{stream()} that returns a \texttt{java.util.Stream} object covering all the elements of the collection. Using Java 8 Streams, the example above can be rewritten in a much more concise and readable manner, as shown in Program 3. 

Program 3 uses the intermediate operation \texttt{filter} that accepts a \texttt{predicate} (a function that takes a given object and returns a true/false value) and produces a new stream object which only covers those elements that satisfy the given predicate (it discards those elements which do not satisfy the predicate). In this particular example, the \texttt{predicate} is defined by the lambda expressions. Next, the terminal operation \texttt{count()} is called on the result to return the number of elements in the filtered stream. As can be seen, using streams and lambda expressions, we can replace with two lines of code the following elements: one class, one interface, three static methods, and two anonymous classes. 

\begin{lstlisting}[title= Program 3 - Use of streams in Java 8, label = Prog3] 
List<Integer> list= Arrays.asList(0,3,5,6,8,10,12,15);
System.out.println("The count of even numbers is" + list.stream().filter(i -> i % 2 == 0).count());
System.out.println("The count of odd numbers is" + list.stream().filter(i -> i % 2 == 1).count());
\end{lstlisting}

\subsubsection{\textbf{Parallel Streams}}\label{sec:java8:parallelStreams}
Let us imagine that our running example involves processing a very large sequence of numbers. It is straightforward to see that this problem can benefit from parallel processing on multi-core processors: the problem can be independently divided into sub-problems (that is, counting even/odd numbers for non-overlapping sub-chains) which can be solved simultaneously  (in parallel, such that each sub-chain is iterated in a separate thread/core), and finally combining the different sub-solutions to obtain a global solution (summing up all the partial counts). 

Java 8 enables coding of this parallel task in a trivial and effortless manner because all the stream operators can either be applied sequentially (as described above, on a single thread) or in parallel (where multiple threads are used). Parallel streams are invoked using the \texttt{parallelStream()} method, present in all the Java collections. Program 4 shows a running example of using parallel streams in Java 8. There is not a need to mention that coding this parallel operation using older Java versions might require a far more complex, hard to read and debug, Java code.

\begin{lstlisting}[title= Program 4 - Use of parallel streams in Java 8, label = Prog4] 
List<Integer> list = Arrays.asList(0,3,5,6,8,10,12,15);
System.out.println("The count of even numbers is" + 
                                             	list.parallelStream().filter(i -> i % 2 == 0).count());
System.out.println("The count of odd numbers is" + 
                                             	list.parallelStream().filter(i -> i % 2 == 1).count());
\end{lstlisting}

\subsection{Bayesian networks}\label{sec:bn}

\begin{figure}[ht!]
\begin{center}
\includegraphics[scale=1]{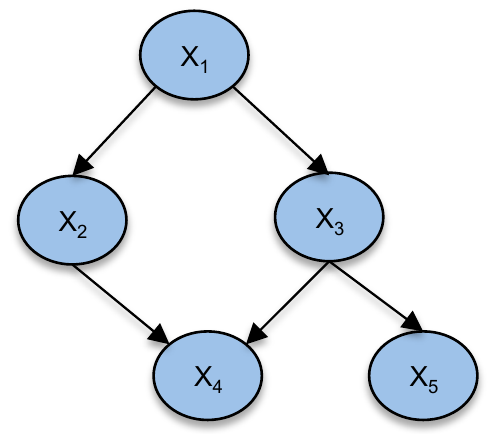}
\caption{\label{fig:ExampleBN} Example of a BN model with five variables.}
\end{center}
\end{figure}

Bayesian networks (BNs) \cite{JensenNielsen07,Pearl88,lauritzen1996graphical} are widely used PGMs for reasoning under
uncertainty.  A BN defined over a set of random variables $\mathbf{X} = \{X_1,\ldots,X_n\}$ can be graphically represented by a directed acyclic graph (DAG). Each node, labelled $X_i$ in the example graph depicted in Fig. \ref{fig:ExampleBN}, is associated with a factor or conditional probability table $p(X_i|Pa(X_i))$, where $Pa(X_i)$ are the parents of $X_i$ in the graph. A BN representation defines a \emph{joint probability distribution} $p(\mathbf{X})$ over the variables involved which factorizes, according to the chain rule, as follows 

\vspace{-0.15in}
$$ p(\mathbf{X}) = \prod_{i=1}^n p(X_i|Pa(X_i)).$$ 

Fig. \ref{fig:ExampleBN} shows an example of a BN model over five variables. A conditional
probability distribution is associated with each node in the network describing its conditional probability
distribution given its parents in the network, and the joint
distribution factorises as  $p(X_1,\ldots,X_5) = p(X_1) p(X_2|X_1) p(X_3|X_1) p(X_4|X_2,X_3) p(X_5|X_3).$

\vspace{0.2in}
\vspace{0.2in}

A BN is called \emph{hybrid} if some of its variables are discrete while others are continuous. In the AMIDST modelling framework, we specifically consider \emph{conditional linear Gaussian (CLG) BNs} \cite{Lauritzen1992,lauritzen1996graphical,LauritzenJensen2001}. In a CLG model, the local distributions of the continuous variables are specified as CLG distributions and the discrete variables are required to only have discrete parents.



\section{Designing Data Structures for PGMs using Java 8}\label{sec:dataStructures}

One of the main conclusions drawn when developing the AMIDST toolbox is that the design of the basic data structures for PGMs should be carefully set up to enable the use of a functional programming style when developing the subsequent algorithms. As we discuss in this section, many different software design proposals could eventually be found for representing the same data structure, but not all of them are valid for later use by the functional programming methods of Java 8 API. 

The AMIDST toolbox contains two main distinct data structures for representing a PGM: a first data structure to represent the directed acyclic graph (DAG) defining the graphical structure of the model; and a second data structure that stores the parameters of the conditional probability distributions. 
 
Many different implementations of a DAG can be imagined and, probably, all of them will have their pros and cons. For example, we can represent a DAG by a boolean matrix, as in the case of the well-known BNT package \cite{murphy2001bayes}, where the $(i,j)$-entry is set to $true$ to represent a directed link from the $i$-th random variable to the $j$-th random variable. In this section, we simply discuss our experiences with those different representations that later on we found to be more suitable for being integrated with the Java 8 API. Our main lesson learnt and our suggested \textit{design pattern} \cite{gamma1994design} is the following: represent data structures as lists of self-contained and independent elements. We exemplify this intuition by discussing the representation of two of our main data structures. 

In the AMIDST toolbox, a DAG is represented as a list of \texttt{ParentSet} objects, one for each variable defining our PGM model (see Section \ref{sec:bn}). Each \texttt{ParentSet} object is self-contained by keeping the following information: a field pointing to the variable stored in this vertex of the DAG; and a list of the parent variables of this vertex. As an example, in Program 5, we show how our DAG representation allows us to concisely code a parallel version of two standard operations for DAG objects, namely, \texttt{getNumberOfLinks()} and \texttt{getChildrenOf(Variable mainVar)}, that can be quite beneficial in terms of computational performance for very large DAGs. Using the aforementioned alternative matrix-based representation for DAGs, it would not be possible to code these methods in such a concise and elegant manner. 

The method \texttt{getNumberOfLinks()} in Program 5 simply iterates over the list of \texttt{ParentSet} objects defining the DAG using a Java \texttt{Stream} object. Then, using the \texttt{mapToInt} operator, it transforms each \texttt{ParentSet} object into an integer value representing the number of parent variables of this object. The returned \texttt{Stream} of integer values is finally reduced with the terminal operator \texttt{count}.

The method \texttt{getChildrenOf(Variable mainVar)} in Program 5 also iterates over the list of \texttt{ParentSet} objects. Then, it filters those \texttt{ParentSet} objects which contain the variable \texttt{mainVar} as a parent variable. Using the \texttt{map} operator, it subsequently extracts for each filtered \texttt{ParentSet} object its main variable. Finally, it collects all these main variables in a list using the terminal operator \texttt{collect}.

As mentioned before, the subtype of PGMs considered and implemented in the AMIDST toolbox is BNs, see Section \ref{sec:bn}. Once again, we found that the best way to represent the data structure for the parameters of a BN was to use a list of \texttt{ConditionalDistribution} objects. Each \texttt{ConditionalDistribution} object is self-contained by keeping, among other things, a reference to all the variables involved (child and parent variables) and a method to compute the local sufficient statistics vector from a given data sample \cite{WinnBishop2005}.

\begin{lstlisting}[title= Program 5 - Determine the number of links in a given DAG \& the set of children of a given variable, label = Prog5]
// This method efficiently calculates the number of links in a particular DAG (represented by this) 
public long getNumberOfLinks(){
	return this.getParentSets().parallelStream()
                    .mapToInt(parentSet -> parentSet.getParents().size()).count();
}
// This method efficiently returns all the children for a particular Variable
public List<Variable> getChildrenOf(Variable mainVar){
	return this.getParentSets().parallelStream()
				      .filter(parentSet -> parentSet.getParents().contains(mainVar))
				      .map(parentSet -> parentSet.getMainVar())
				      .collect(Collectors.toList());
}
\end{lstlisting}
\vspace{0.1in} 

A standard task involving this data structure is to compute the global sufficient statistics vector for a BN model. By definition, this global sufficient statistics vector is composed of the concatenation of all the local sufficient statistics vectors associated with each \texttt{ConditionalDistribution} object of the BN. Program 6 shows the code for computing this ``compound'' vector. It starts by iterating over the \texttt{ConditionalDistribution} objects of a BN and mapping each one with a new created \texttt{IndexedElement} object. Each \texttt{IndexedElement} object contains the local sufficient statistics vector associated to the mapped \texttt{ConditionalDistribution} object and it is indexed by the corresponding ID of its main variable (that is why it was important to include this variable in the data structure and make it self-contained). Finally, all these newly created \texttt{IndexedElement} objects are collected in a list through the terminal operator \texttt{Collector}. 

\vspace{0.1in}
\begin{lstlisting}[title= Program 6 - Compute a BN global sufficient statistics vector, label = Prog6] 
// This method returns an instantiation of our defined interface Vector that represents the global vector of sufficient statistics of a BN for a given DataInstance. In order to efficiently accomplish that, both ConditionalDistribution and IndexedElement objects are defined as independent and self-contained
public Vector getSufficientStatistics(DataInstance dataInstance){
	List<IndexedElement> list = bayesianNetwork.getConditionalDistributions()
					.stream()
					.map(condProb -> 
						new IndexedElement(condProb.getVariable().getIndex(), 
								condProb.getSufficientStatistics(dataSample))
					.collect(Collectors.toList());
	return new CompoundVector(list);
}
\end{lstlisting}
\vspace{0.1in} 

As can be seen, the final output of this method is a \textit{CompoundVector} object, which implements our defined interface \textit{Vector}. This  \textit{CompoundVector} is again designed as a list of independent and self-contained \textit{IndexedElement} objects (desired properties for parallelisation). This \textit{CompoundVector} class is frequently used in AMIDST as it allows us to parallelise several vector operations. 



\section{Parallelisation of algorithms}\label{sec:algorithms}

As commented on in previous sections, Streams in Java 8 allow us to parallelise tasks without having to explicitly deal with threads and their synchronisation. Note that, sometimes, not all the tasks should be parallelised, since an attempt to do so may result into more inefficient versions comparing to their sequential counterparts. Here we include a non-exhaustive list of the main criteria that an algorithm should fulfil in order to be potentially parallelisable when implemented in Java 8:

\begin{itemize}

\item It is generally desired that all parallel subtasks, into which the algorithm is decomposed, entail a significant CPU-intensive processing (on the order of 1-10 milliseconds). More specifically, this processing time has to be large enough to compensate for the overload that may be introduced due to both thread creation (and synchronisation if needed) and I/O operations for the case of parallel data processing. 

\item All parallel subtasks must run completely independently or only access a non-modifiable shared data structure. This stateless property is actually a requirement for the use of parallel Java Streams \cite{urma2014java}.

\item The low-latency random access of the elements of the parallel stream represents an additional requirement. In Section \ref{sec:dataStreams}, we discuss what to do in cases where this requirement is not readily available, as it occurs when the elements are placed on a hard disk drive.

\end{itemize}

We exemplify all these insights through three different algorithms for PGMs, namely, parallel maximum likelihood estimation, a parallel Monte Carlo method using importance sampling, and a greedy search method for solving a combinatorial optimisation problem also performed in parallel. Many other parallel algorithms are also included in our AMIDST toolbox. As mentioned before, a key advantage of the new Java 8 API is that it facilitates coding parallel algorithms using just a few lines of code. 

\subsection{Parallel maximum likelihood estimation}\label{sec:parallelml}

Maximum likelihood estimation (MLE) \cite{mlestimation} is an estimation method of probability functions. Given a particular data set of i.i.d. samples $D = \{\bmx_1,\ldots, \bmx_n\}$ and an underlying statistical model, which in our case is a BN whose probability distributions belong to the exponential family \cite{WinnBishop2005}, the MLE method gives a unified approach for calculating the parameters $\boldsymbol{\theta}$ of the model that maximize the logarithm of the likelihood function (log-likelihood function) as follows 

\begin{equation}
\label{eq:ml}
\boldsymbol{\theta}_{MLE} = \argmax \sum_{i=1}^{n}\log p(\bmx_i|\boldsymbol{\theta}) = \frac{\sum \mathbf{s}(\bmx_i)}{n},
\end{equation}

\noindent where $\mathbf{s}(\cdot)$ is a deterministic function that returns, for each data sample, a vector of sufficient statistics, which depends on the particular model we are learning.  Note that, the previous equation only holds for fully observed data sets and models that do not contain latent variables. 

The key point is that the MLE algorithm can be expressed as the computation of an independent sum operation over a collection of data samples, and satisfies the three aforementioned parallelisation requirements. The MLE computations can therefore be performed in an embarrassingly parallel manner. In what follows, we show in Program 7 how MLE can be implemented in Java 8 using just a few lines of code. 

\begin{lstlisting}[title= Program 7 - Parallel maximum likelihood estimation, label = Prog7] 
// This method computes the MLE for a given DataStream and a BayesianNetwork
// The input parameter batchSize can be used to customise the degree of parallelisation 
SufficientStatistics computeMLE(DataStream data, BayesianNetwork bayesianNetwork){
	SufficientStatistics sumSS = data.parallelStream(batchSize)        
					     .map(bayesianNetwork::getSufficientStatistics) //see Program 6
					     .reduce(new ZeroVector(), Vector::sumVector);
	return sumSS.divideBy(data.getNumberOfInstances());
}
\end{lstlisting}
\vspace{0.15in}

\begin{figure}[ht!]
\begin{center}
\includegraphics[scale=0.65]{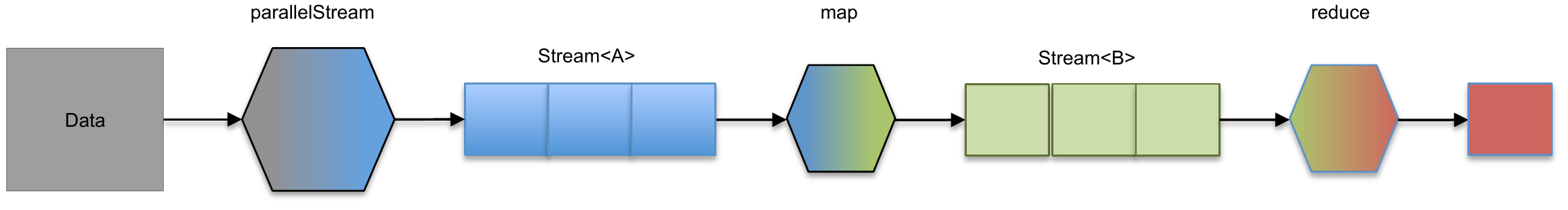}
\caption{\label{fig:mapreduce} Processing flow of the maximum likelihood estimation algorithm detailed in Program 7.}
\end{center}
\end{figure}

The MLE algorithm is based on \texttt{map} and \texttt{reduce} operations. An illustration of this processing flow is given in Fig. \ref{fig:mapreduce}. Our data set is accessed through the \texttt{data} object. The method \texttt{parellelStream(int batchSize)} produces a stream of \texttt{DataInstance} objects. The \texttt{map} operator transforms the stream of \texttt{DataInstance} objects to a stream of \texttt{Vector} objects. This transformation is made using the \texttt{bayesianNetwork::getSufficientStatistics} method previously described in Program 6 which, given a \texttt{DataInstance} object, produces a \texttt{Vector} object (it implements the function $\bms(\cdot)$ referenced in Equation \ref{eq:ml}). Finally, the stream of \texttt{Vector} objects containing all the sufficient statistics is reduced to a single \texttt{Vector} object using the reduction operation passed as an argument to the \texttt{reduce} operator. This reduction operation is \texttt{Vector::sumVector}, that is, a method that returns the sum of two given vectors. The first input argument of the \texttt{reduce} operator is \texttt{new ZeroVector()} which represents the final accumulator object. 

\vspace{0.35in}
\vspace{0.35in}

The key element of this parallel MLE algorithm is the use of a stream of fixed-sized data batches created by invoking the method \texttt{parallelStream(batchSize)}. This method is coded using the functionality described in Section \ref{sec:dataStreams}. As will be shown in Section \ref{sec:experimentalEvaluation}, the size of the batch selected plays an important role in the parallelisation performance. 

Again, this example illustrates how simple it is to code a parallel version of an algorithm using Java 8. Even though, traditionally, it has not been trivial to parallelise this widely used parameter estimation method in practice. For instance, in 2002, Swann \cite{mleMPISwann2002}  presented a 34-page paper to lower the cost of using MPI to solve a maximum likelihood problem. 

\subsection{Parallel Monte Carlo}\label{sec:parallelmc}

Monte Carlo \cite{hammersley1964monte} methods are a widely known and computationally intensive family of algorithms that has been heavily used in the last few decades for solving a broad class of physical, mathematical, and statistical problems. Many Monte Carlo methods can be efficiently run by resorting to parallel implementations. However, similar to the MLE algorithm, the support for parallel implementations of Monte Carlo methods in PGM toolboxes is quite limited \cite{tristan2014augur}. 

In this section, we show how by using Java 8, we can easily implement in a few lines of code parallel versions of a relevant type of a Monte Carlo method: the Importance Sampling (IS)\cite{hammersley1964monte} algorithm. IS can be defined as a variance reduction Monte Carlo technique. It is based on producing samples, using an auxiliary probability distribution, from interesting or important regions of our probability distribution.  For instance, in many applications, we want to calculate $\mathds{E}(f(\mathbf{X}))$, that is, the expected value of a function of interest $f(x)$ according to some probability distribution $p$. However, for many interesting applications in Bayesian inference \cite{hammersley1964monte}, this expected value cannot be computed exactly (usually because $p$ is intractable). IS algorithms compute the estimates of this expected value by generating a large set of samples $\{x_i\}_{i \in \{1,\ldots,m\}}$ from an auxiliary tractable sampling distribution $q$ that overweights the interesting region, and then adjusts the estimation to account for the oversampling from this auxiliary distribution with the weight $w = p(x_i)/q(x_i)$ for each sample $x_i$, $i \in \{1,\ldots,m\}$. The estimate of the expected value is computed as follows,

\vspace{-0.5cm}
\begin{equation}\label{eq:importancesampling}
\mathds{\hat{E}}(f(\mathbf{X})) = \frac{\sum_{i=1}^m f(x_i)w_i}{\sum_{i=1}^m w_i}\cdot
\end{equation}

Hence, an IS algorithm can be decomposed into the computation of two different sums and fulfils the aforementioned parallelisation requirements. The Java 8 code of the IS algorithm is shown in Program 8.1. In the first part of the code, we randomly generate a set of samples $\{x_i\}_{i \in \{1,\ldots,m\}}$ and store them in a list, the \texttt{weightedSampleList} object. This sampling process can be trivially parallelised using Java 8 Streams. A parallel stream of \texttt{Integer} objects is created with the desired \texttt{sampleSize}, which allows generating weighted samples in parallel via the operator \texttt{mapToObj}. Note that the \texttt{samplingModel} object already contains information about the sampling distribution. In the second part of this code, we iterate over the previously generated samples to compute the expected value of the function $f$ according to Equation \ref{eq:importancesampling}. More precisely, each of the previously generated samples is mapped to a two-element list, where the first element corresponds to the sum of the numerator and the second element corresponds to the sum of the denominator of Equation \ref{eq:importancesampling}. The \texttt{reduce} operator simply adds up the two-element list in a similar way as it was used in Program 7. The code snippet in Program 8.2 shows an example of two queries that can be invoked, and illustrates also the use of lambda expressions.

Note that there is a subtle issue in Program 8.1 which needs further discussion. In this code, we have used the class \texttt{java.util.Random} to generate, in parallel, pseudo-random numbers in order to produce the weighted samples. Instances of \texttt{java.util.Random} are thread-safe. However, the concurrent use of the same \texttt{java.util.Random} object across threads will encounter contention and may potentially lead to a poor performance. Instead, we could use the class \texttt{ThreadLocalRandom} in a multithreaded environment, in order to generate random numbers locally in each of the threads running in parallel, as shown in Program 8.3.

\vspace{0.1in}
\begin{lstlisting}[title= Program 8.1 - Parallel importance sampling, label = Prog81] 
public double getExpectedValue(String varName, Function<Double,Double> f){
	Random random = new Random(0);
	// 1. Parallel random generation of samples
	Stream<WeightedInstance> weightedSampleList = 
						IntStream.range(0, sampleSize).parallel()
						.mapToObj(i -> samplingModel.getWeightedSample(random))
	// 2. map: each sample is mapped into two elements
	// 3. reduce: numerators and denominators for all samples are summed up
	List<Double> sum = 
			weightedSampleList.parallel()
			.map(w-> Arrays.asList(f(w.getValue(varName)) * w.getWeight(), w.getWeight()))
			.reduce(Arrays.asList(0.0,0.0), 
				(a,b)-> Arrays.asList(a.get(0)+ b.get(0), a.get(1)+ b.get(1)));
	// 4. The resulting values refer to the numerator and denominator in Equation 2, respectively
	return sum.get(1)/sum.get(0);
}
\end{lstlisting}
\vspace{0.1in}

\begin{lstlisting}[title= Program 8.2 - Examples of queries using Program 8.1, label = Prog82] 
System.out.println("p(A<0.7) = " + ImportanceSampling.getExpectedValue("A", a->{(a<7)? 1.0 : 0.0}));
System.out.println("E(A) = " + ImportanceSampling.getExpectedValue("B", b->b);
\end{lstlisting}
\vspace{0.1in}

\begin{lstlisting}[title= Program 8.3 - Use of ThreadLocalRandom for generating multithreaded random numbers (internally generated seed), label = Prog83] 
	.mapToObj(i -> SamplingModel.getWeightedSample(ThreadLocalRandom.current()))
\end{lstlisting}
\vspace{0.2in}

Although we are avoiding the collision of threads when accessing a shared resource, \texttt{ThreadLocalRandom} objects are initialised with an internally generated seed that cannot otherwise be modified using the current Java API, so the results are not reproducible. Thus, we introduce in Program 8.4 another alternative for parallel random number generation. We create our own \texttt{LocalRandomGenerator} that will allow us to generate pseudo-random numbers locally in each thread with a particular seed. This approach might lead to some overhead, by creating new instances of Random objects. However, it allows us to provide a reproducible alternative.

In \cite{CAEPIA2015} we provide an experimental analysis that shows high computational performance improvements with the parallelisation of IS algorithms based on the presented design solutions using the AMIDST toolbox.

\begin{lstlisting}[title= Program 8.4 - A solution for generating parallel random numbers, label = Prog84] 
// Class definition for reproducible multithreaded random number generation
public class LocalRandomGenerator {
	final Random random;
	public LocalRandomGenerator(int seed){
		random = new Random(seed);
	}
	public Random current(){
		return new Random(random.nextInt());
	}
}
// Example of use
LocalRandomGenerator randomGenerator = new LocalRandomGenerator(seed);
...
  .mapToObj(i -> SamplingModel.getWeightedSample(randomGenerator.current()))
...
\end{lstlisting}
\vspace{0.1in}
\subsection{Parallel greedy search}\label{sec:parallelgs}

There are many different domains in which a greedy search (GS) algorithm can be used as a heuristic to find a local optimal solution to a combinatorial optimisation problem. When more exhaustive algorithms are computationally prohibitive, then a local solution can be a viable approximation to the global one. For PGMs, GS algorithms are widely used to solve problems such as structural learning \cite{heckerman1995learning} or wrapper feature subset selection \cite{kohavi1997wrappers}.

\begin{lstlisting}[title= Program 9.1 - Parallel greedy search for wrapper feature subset selection, label = Prog91] 
// WrapperFeatureSelection implements the GreedyAlgorithm interface
GreedyAlgorithm<Variable> algorithm = new WrapperFeatureSelection(data, classVariable);
Candidate<Variable> bestCandidate = algorithm.createEmptyCandidate();
bestCandidate.setScore(Double.NEGATIVE_INFINITY);
// Sequential part: compare the best candidate for each iteration against the global best candidate
do{
	List<Candidate<Variable>> candidates = algorithm.getCandidates();
	//Parallel part: get the best candidate for the current iteration (inner candidate)
	Candidate<Variable> bestInnerCandidate = candidates.parallelStream()
		                     .map(candidate -> algorithm.evaluate(candidate))
		                     .reduce(bestCandidate, 
		                     	(a, b) -> {if(a.getScore() > b.getScore() + threshold) 
		                     			return a; 
		                     		   else 
		                    			return b;
		                    	});
	algorithm.updateBestCandidate(bestCandidate);
}while(bestInnerCandidate != bestCandidate);
\end{lstlisting}

In a very general context, a GS algorithm requires 1) a set of candidates, from which a solution is gradually created; 2) an objective function, which evaluates the fitness of a candidate solution; 3) a selection function, which chooses the best candidate to add to the solution; and 4) a solution function, which indicates if the current solution is the final one. 

Contrary to the two above-described MLE and IS algorithms, a GS algorithm cannot be fully parallelised. In fact, a GS algorithm is sequential, which means that the solution of the next step depends on the solution of the current step. However, there is room for parallelisation when building the candidate solution at a specific step. 

For instance, in many cases, the selection function consists simply of finding the element in a very large set of candidates with the highest score according to the objective function. Program 9.1 contains the Java 8 code of a general parallel GS algorithm (particularised for wrapper feature subset selection \cite{kohavi1997wrappers}). In each iteration of the main do-while loop, the \texttt{bestInnerCandidate} is evaluated and selected in parallel using \textit{parallel streams}, \texttt{map}, and \texttt{reduce} operators (in this case, the \texttt{reduce} operator finds the element in the stream with the maximum value). This code can be instantiated into different greedy search problems by implementing the \textit{Candidate} and \textit{GreedyAlgorithm} interfaces defined in Program 9.2. 

Again, to the best our knowledge, there is no PGM toolbox which provides this kind of functionality. The above example shows how the implementation of a parallel greedy search solver is quite straightforward when using Java 8 API.

\begin{lstlisting}[title= Program 9.2 - Interfaces to be implemented for any greedy search problem, label = Prog92] 
public interface Candidate<E> {
	E getObject();
	double getScore();
	void setScore(double score);
}
public interface GreedyAlgorithm<E> {
	Candidate<E> createEmptyCandidate();
	List<Candidate<E>> getCandidates();
	Candidate<E> evaluate(Candidate<E> candidate);
	void updateBestCandidate(Candidate<E> candidate);
}
\end{lstlisting}
\vspace{0.1in}



\section{Parallel processing of data sets}\label{sec:dataStreams}

In many settings, data sets can be defined as a collection of independent items (documents, customer profiles, genetic profiles, patient health records, etc.), and thereby many data mining and machine learning techniques can be described as algorithms which independently process each of these items, once or multiple times. For instance, this applies to any algorithm fitting the statistical query model \cite{kearns1998efficient,chu2007map} or any Bayesian learning approach over i.i.d. data samples and exponential family models \cite{broderick2013streaming}, as in the case of the MLE algorithm discussed in Section \ref{sec:parallelml}.

The latter settings are quite suitable for coding parallel implementations of these algorithms using Java 8 Streams. When developing the AMIDST toolbox we heavily used the  following design pattern for many data processing algorithms by creating a stream of data item objects. Each data item is processed using a given operation according to the specific algorithm (usually using a \texttt{map} operation) and all intermediate results are combined using a particular reduction operation. The parallel MLE algorithm proposed in Section \ref{sec:parallelml} is an example of these kinds of algorithms and many other machine learning methods can be expressed in terms of map-reduce operations \cite{chu2007map}.

This parallelisation approach based on Java 8 Streams needs, in principle, the data set to be stored in RAM to support parallel random access to the different data samples, as discussed in Section \ref{sec:java8:parallelStreams}. Although the RAM allocation of the data set is doable in many cases, modern data sets are usually too big to be held in RAM. When  data is placed on a hard disk, it has to be accessed in a purely sequential manner, which is in complete contradiction with the random access requirements of parallel processing. 

In the next subsection, we discuss the design solution used in AMIDST to approach this problem and set the basis for the development of parallel algorithms for processing large data samples that do not fit in main memory. In AMIDST, we employ this functionality for many different algorithms, as it is the case for the previously discussed MLE algorithm (see Section \ref{sec:parallelml}). We also use this solution in AMIDST to provide, for example, a parallel implementation of the \textit{K-means} algorithm for data sets that do not fit in memory, which is coded using a few lines of code (plus some simple external classes).  If we look at other parallel implementations of the \textit{K-means} algorithm, for example, using the OpenMP \cite{tian2002intel} framework in C++\footnote{\webOpenMP}, we see that it is far more complex to code and requires explicit multi-thread handling (as well as it assumes that all data fit in main memory!). 

\subsection{Streams of data batches}\label{sec:streamDataBatches}

\begin{figure}[ht!]
\begin{center}
\includegraphics[scale=0.9]{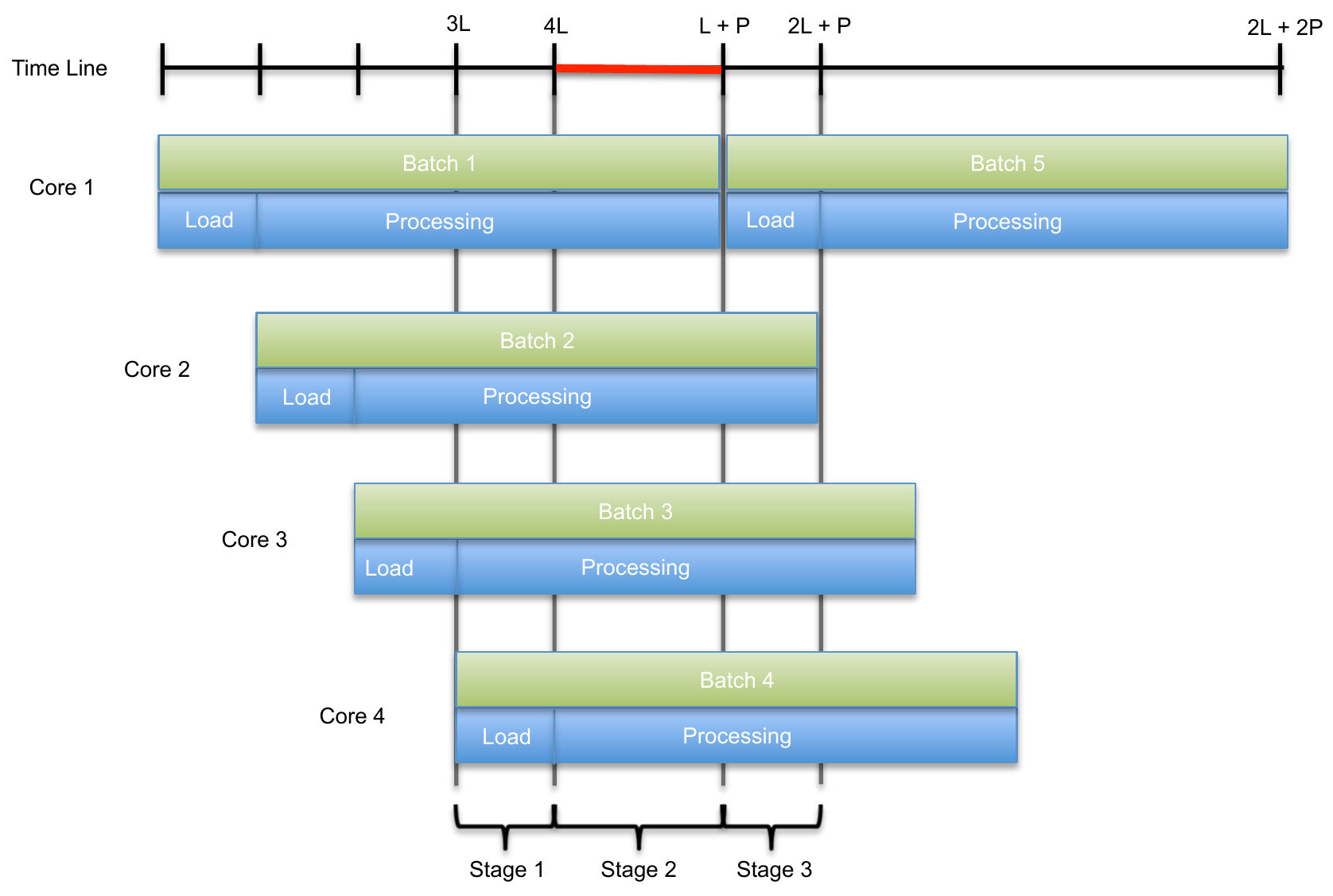}
\caption{Example of the distribution of five data batches with the same demands in terms of loading and processing time, denoted as $L$ and $P$ respectively, in a four-core CPU. The red line indicates the time frame for which all data batches are being processed at the same time.\label{fig:dataStream}}
\end{center}
\end{figure}

In this section, we describe a solution for the above problem based on the use of streams of data \textit{batches} (a small group of data items). This approach loads data batches on demand, then performs parallel processing of the data batches by exploiting the fact that, for many machine learning algorithms, the time of processing a data batch, denoted by $P$, is larger than the time to load this batch from the hard disk to main memory, denoted by $L$. Using this approach, the first core loads the first batch of data. Next, this first core starts to process it, releases the I/O disk channel, and meanwhile, the second core starts loading the second batch of data. In Fig. \ref{fig:dataStream}, we visually illustrate how this standard approach would work on a four-core CPU for a setting where $P>3*L$ (the time of processing a data batch is 3 times larger than the time for loading this batch from disk). In the case of Java 8, this could be performed by relying on the lazy evaluation feature of Java 8 Streams (see Section \ref{sec:java8:streams}).

It is not hard to see that this strategy is limited in the level of parallelism it can achieve, which basically depends on the ratio between $P$ and $L$. More precisely, with this approach, the maximum number of cores working at the same time is $\lfloor P/L +1 \rfloor$. This quotient depends on the algorithm and the CPU computing power, through the value of $P$, as well as the I/O speed, thorough the value of $L$. Moreover, it also depends on the size of the batch that affects both $P$ and $L$ and represents the unique parameter that can be tuned in order to achieve an optimal performance. 

\vspace{0.2in}
\vspace{0.25in}

\subsection{The \textit{Spliterator} interface}

Unfortunately, Java 8 does not provide native methods for creating a stream of batches of the same size from a disk file\footnote{We could use the class \textit{java.nio.file.Files} to obtain a parallel stream of lines from a file but the size of the batches can not be set and grows in arithmetic progression.}. However, we can create customized Java Streams by using the \textit{java.util.Spliterator} interface\footnote{See \url{https://www.airpair.com/java/posts/parallel-processing-of-io-based-data-with-java-streams/} for further details on this approach.}. A very important property of this interface is that we do not need to create \textit{thread-safe} classes implementing this interface. Java 8 guarantees to use your \textit{Spliterator} objects in a single thread at a time. This makes the implementation of \textit{Spliterator} a more simple operation. 

In what follows, we show which are the main elements of the \textit{Spliterator} object, providing a stream of batches of a fixed size. Our \textit{Spliterator} implementation assumes that the data set is placed in a text file where each line of the file contains a data item (usually a data item is composed of a sequence of comma-separated numbers, but it can also be, for example, a long string containing an XML document). \textit{Spliterator} is a combined name which refers to two main abstractions behind Java Streams: \textit{iterating} and \textit{splitting}. These two abstractions are specified in the following two methods of the \textit{Spliterator} interface:

\begin{itemize}
\item \verb+boolean tryAdvance(Consumer<? extends T> action)+: This method performs the actual processing of the stream. It consumes one item of the stream and invokes the passed-in \textit{action} on it. As shown in Program 10, this method is implemented as follows: it first reads a new line from the file, creates a new data item object from this line, and then processes it with the action object. The action object encapsulates the operation that will be applied later to the data item in the processing flow of the Java Stream by using the methods \textit{map}, \textit{filter}, etc.

\begin{lstlisting}[title= Program 10 - Iterating abstraction of the Spliterator, label = Prog10] 
// This method performs the actual processing of the stream
boolean tryAdvance(Consumer<DataItem> action){
	// 1. Read one line
	String line = getBufferedReader().readLine();
	if (line!=null){
		// 2. Create a new DataItem object from each line
		DataItem dataItem = new DataItem(line);
		// 3. Process the DataItem with the action object
		action.accept(dataItem);
		return true;
	}else{ return false; }
}
\end{lstlisting}
\vspace{0.3in}
 
\item \verb+Spliterator<T> trySplit()+: This method is in charge of splitting off a part of the sequence that the current \textit{Spliterator} is handling. The result is a new \textit{Spliterator} object that will handle the new split sequence. Java 8 uses this method to perform the parallel processing of the stream. At the beginning, when a stream operation is invoked over a list of elements, a \textit{Spliterator} object is created and handled by one thread. Then, at some point, this thread invokes the \texttt{trySplit} method to create a new \textit{Spliterator} object and assign it to a new thread. That new object will now be in charge of processing, in parallel, the new split sequence. In our case, when this method is invoked, we read $B$ consecutive lines, with $B$ being the size of the data batch, and create a new \textit{Spliterator} object for handling this sequence of $B$ lines. In that way, once a new thread is assigned to this new \textit{Spliterator}, it will only process this batch of data. In Program 11, we detail the Java code for this method in our \textit{Spliterator} class.

\end{itemize} 

Once the \texttt{Spliterator} class is defined, we can create a fixed-batch-size stream using the class java.util.stream.StreamSupport. This is how the method \verb+parallelStream(int batchSize)+ of the class \texttt{DataStream} is coded in AMIDST.  

\vspace{0.4in}
\begin{lstlisting}[title= Program 11 - Spliting abstraction of the Spliterator, label = Prog11] 
// This method split off the part of the sequence for this Spliterator
public Spliterator<DataItem> trySplit(){	
	final HoldingConsumer<DataItem> holder = new HoldingConsumer<>();
	// 1. Check that the end of the sequence is not reached
	if (!spliterator.tryAdvance(holder)) return null;
	// 2. Populate data batch with batchSize items of the sequence
	final DataItem[] batch = new DataItem[batchSize];
	int j = 0;
	do { batch[j] = holder.value; } while (++j < batchSize && tryAdvance(holder));
	// 3. Create a new Spliterator that is assigned to a new thread
	return Spliterators.spliterator(batch, 0, j, characteristics());
}
\end{lstlisting}



\section{Experimental Evaluation}\label{sec:experimentalEvaluation}

\subsection{Sequential vs Parallel Java 8 programming}\label{ss:seqvsparallel}

In this section we give some experimental examples to show the computational advantages of using parallel streams as well as some of the tradeoffs we need to make when coding parallel algorithms in Java 8. For this purpose, we use a simple algorithm like MLE which, as shown in Section \ref{sec:parallelml}, can be trivially parallelised using Java streams.

Ideally, the higher the number of cores is used, the higher the speedup is obtained. But, of course, this speedup is at some point limited by the overload introduced by the thread creation. Moreover, as previously discussed in Section \ref{sec:streamDataBatches}, algorithms based on parallel processing of data batches, as it is the case of parallel MLE, should also consider that there is a limit in the level of parallelisation they can achieve. This limit is determined by the expression $\lfloor P/L +1 \rfloor$, where $P$ denotes the time of processing a data batch and $L$ denotes the time needed to load this batch from the hard disk to main memory. 

\begin{figure}[ht!]
\begin{center}
\includegraphics[scale=0.95]{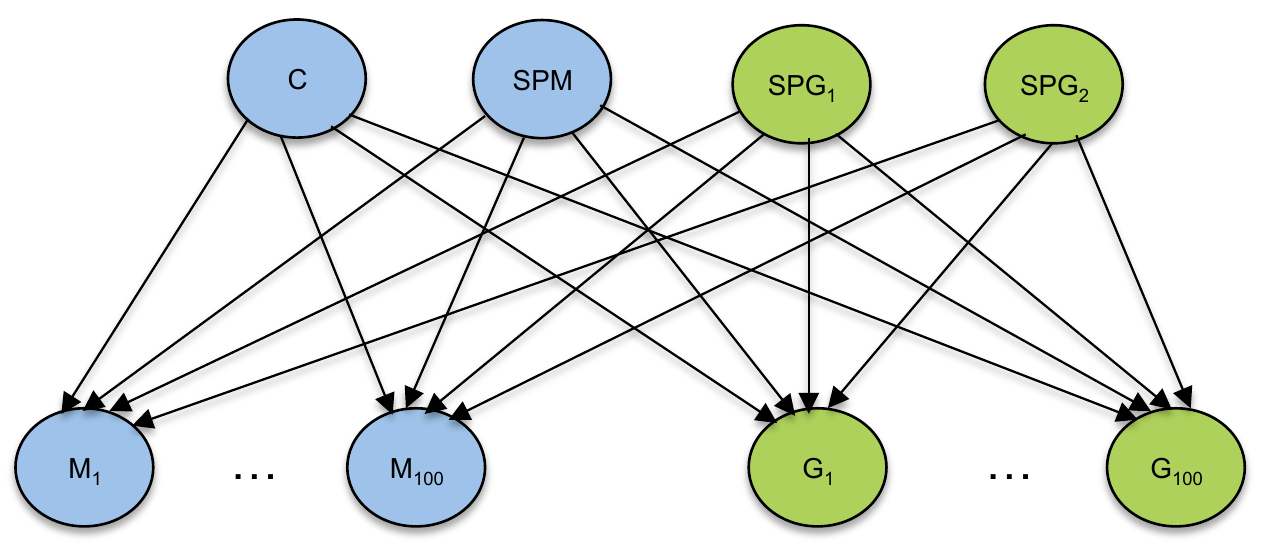} 
\caption{\label{fig:extendedNB} A Bayesian network structure.}
\end{center}
\end{figure}

To show this trade-off, we have conducted some experiments performing MLE using a synthetic data set of one million samples. Samples have been randomly generated using the Bayesian network structure, depicted in Fig. \ref{fig:extendedNB}, with random parameters. The considered Bayesian network consists of two multinomial \textit{super-parent} variables (one of them could be considered the class variable, denoted $C$, and the other denoted $SPM$); two Gaussian \textit{super-parent} variables, denoted $SPG_{1}$ and $SPG_{2}$; and 100 multinomial variables ($M_1, \ldots, M_{100}$) and 100 Gaussian variables ($G_1, \ldots, G_{100}$) that depend on all the super-parent variables. Our task here is to learn the parameters of this Bayesian network using MLE.

\vspace{0.3in}
\vspace{0.3in}

\begin{figure}[ht!]
\begin{center}
\begin{tabular}{cc}
\includegraphics[scale=0.4]{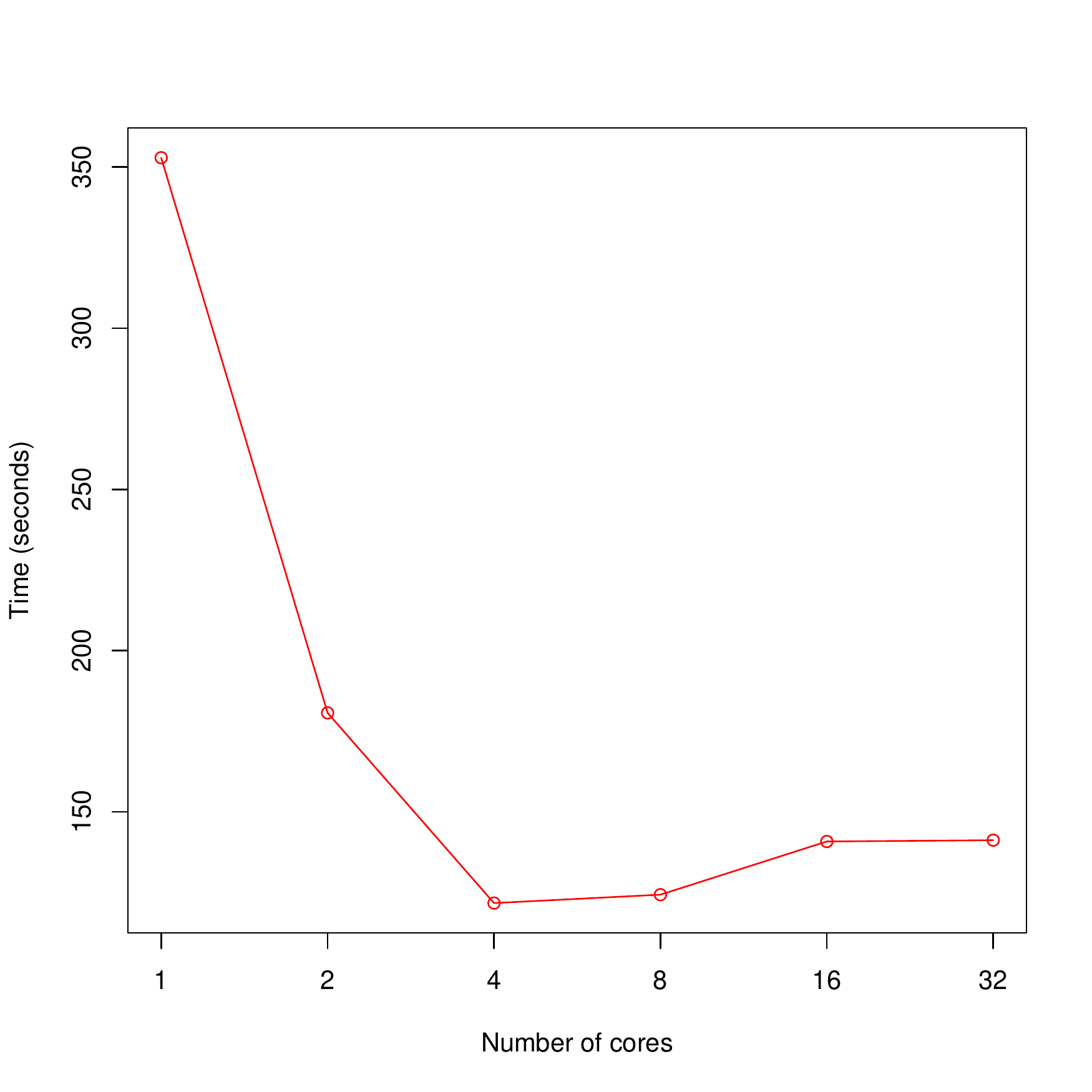} & \includegraphics[scale=0.4]{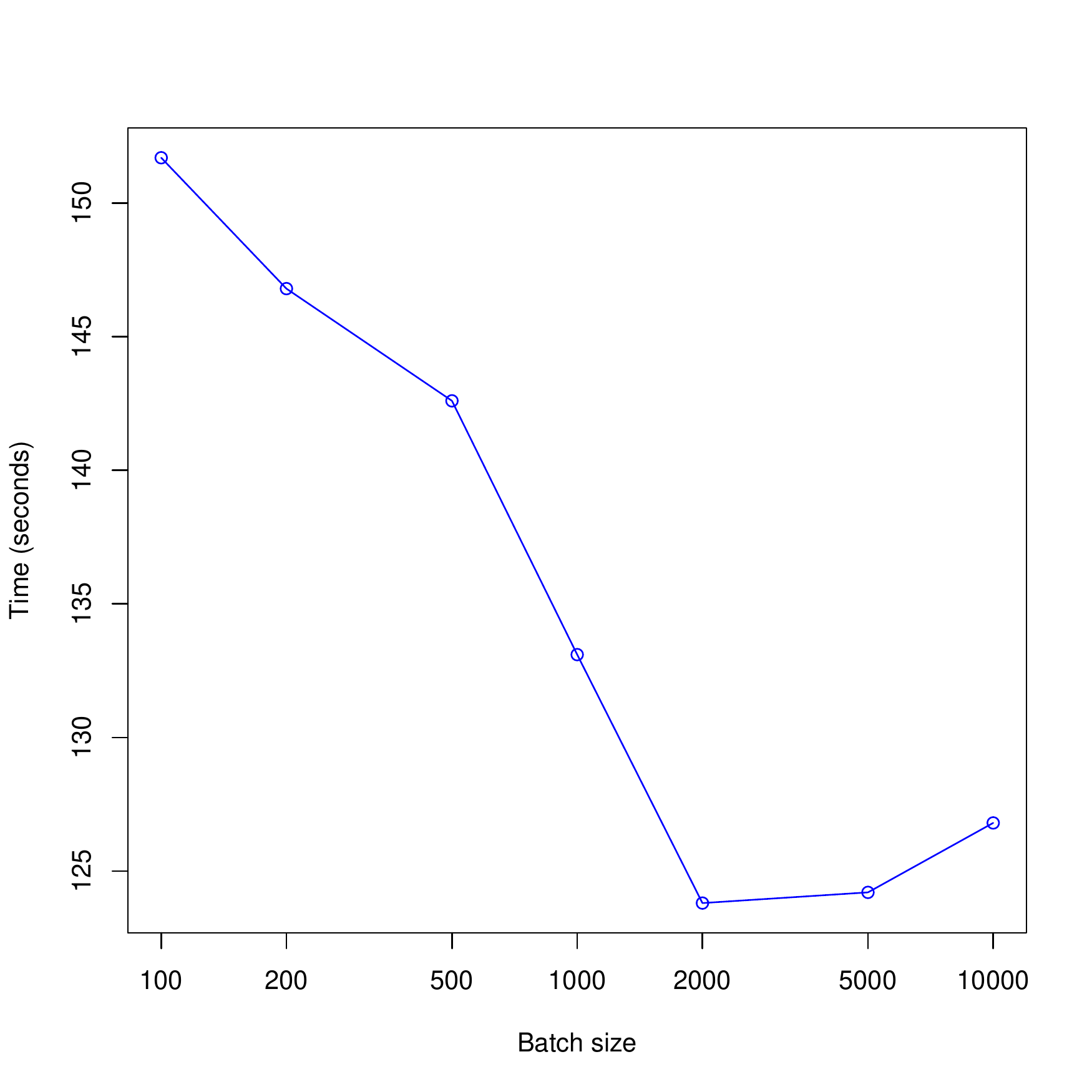}\\
(a) Comparing the number of cores & (b) Comparing the batch size\\
(with batchSize=1000) & (with all available cores)\\
\end{tabular}
\vspace{0.05in}
\caption{\label{fig:mlExperiments} Experiments running the MLE algorithm in parallel for different number of cores and batch sizes.}
\end{center}
\end{figure}

Fig. \ref{fig:mlExperiments}(a) shows the time needed to complete the above task in a machine with 32 cores when fixing the batch size to 1000 samples and varying the number of cores\footnote{Java 8 API does not allow to fix the number of cores to be used in a parallel stream, but we can fix the maximum number of allowed cores by using the linux command \texttt{taskset}, which may introduce some overhead.}. As can be seen, using parallelisation, even with only 2 cores, a significant reduction (by a half) of the time required to learn from a million samples. So, for this simple algorithm, where computational complexity of processing a data batch is linear in the number of samples of the batch, the optimal value is obtained with only four cores. So, as discussed above, it seems that we reach the maximum degree of parallelisation with four cores for a batch size of 1000 samples. By using more cores we introduce some overhead. 

\vspace{0.1in}
\vspace{0.25in}

Fig. \ref{fig:mlExperiments}(b) shows the same parallel MLE experiment carried out in all available 32 cores but modifying the batch size. In this case, the optimal value is obtained with a batch size of $2000$ samples. Using a small batch size introduces a high overload in terms of thread creation because the problem is divided into many small sub-problems. Whereas using a batch size that is too big can underuse the offered capacity of the computer in terms of available cores because the problem is divided into few big sub-problems. In any case, we can observe how these differences in times when using different batch sizes are relatively small (approximately 30 seconds in the worst case) compared to the time differences when no parallelisation is carried out (more than 200 seconds). We believe that the optimal batch size is unfortunately problem dependent. It can be a good idea to perform, for instance, a parallel cross-validation evaluation in order to determine its value. We have so far performed experiments using real datasets of different sizes and observed that the optimal batch size is generally in the order of one thousand \cite{IDA2015,SCAI2015}.

\subsection{Java 8 vs Big data platforms for multi-core parallelisation\label{ss:java8vsspark}}

In this section, we would like to highlight the high overheads associated to a big data platform like Apache Flink \cite{FLINK2015} when used for parallel processing in a multi-core computer. As mentioned before, Apache Flink is an open source platform for distributed stream and batch data processing. It executes arbitrary data-flow programs in a data-parallel and pipelined manner, enabling the execution of bulk/batch and stream processing programs. For this comparison, Apache Flink should behave similar to Apache Spark \cite{Zaharia2010} as they approach similar data distributed processing problems.

For this experiment, we coded in Flink the same parallel MLE algorithm used in the previous section and ran it over the same data set and on the same $32$-cores server. The obtained running time with Flink was $1388$ seconds, that is, around $23$ minutes to learn the parameters of the BN network in Fig. \ref{fig:extendedNB} from 1 million samples. In Java 8, it took less than $3$ minutes to complete the same task.

In this case, Apache Flink's runtime is significantly higher due to the overhead introduced. This overhead is basically related to all the special functionalities covered in order to have a platform that is able to access distributed data files, to recover from failures in the hardware, to balance the computational load across the nodes of the clusters, etc. Therefore, we may conclude that, in general, such big data platforms are especially designed to preferably run on multi-node clusters of commodity hardware, not on a single computer with multi-core processing units.


\section{Conclusions}\label{sec:conclusion}

In this paper, we have presented different design solutions to deal with PGMs on multi-core CPUs using the new Java 8 API. We have also demonstrated how this new API provides a rich set of functionalities that make the coding of concurrent/parallel applications a far simpler task compared to previous Java releases and, probably, to many other programming languages. In addition, we have provided some general guidelines on how to design data structures, algorithms, and parallelisable data processing methods using Java 8 features. We hope that these guidelines can be of help to developers and researchers that aim to build computational intelligence software exploiting multi-core CPU architectures. 

In our opinion, a key aspect that makes the functional programming Java 8 features quite appealing for developing computational intelligence software is that these functional primitives decouple the problem of what can be parallelised from the problem of how to run an algorithm in parallel. Actually, once the algorithm has been expressed in terms of map-reduce operations over Java streams, we cannot control which part of the computations is performed on a specific core or how many cores should be used. This is completely transparent to the user. In that way, we are really going beyond multi-core CPU architectures. For example, researchers at Oracle (the Java owner) \cite{su2014changing} recently proposed an extension of Java Streams, called \texttt{DistributableStream}, for data processing over a cluster of computers with a distributed file system such as Hadoop or Cassandra. With this approach, the same code that runs in parallel on a multi-core CPU using native Java Streams, can also be run on a distributed cluster using the class \texttt{DistributableStream}. We envision that the forthcoming Java releases may give support to other kinds of parallel computing architectures introducing more specifications on what can be parallelisable rather than how to perform parallel computing.

\section*{Acknowledgments}
This work was performed as part of the AMIDST project. AMIDST has received funding from the European Union's Seventh Framework Programme for research, technological development and demonstration under grant agreement no 619209.

\bibliographystyle{IEEEtran}

\bibliography{biblio}

\end{document}